\documentclass[pdflatex,sn-mathphys-num]{sn-jnl}

\usepackage{graphicx}%
\usepackage{multirow}%
\usepackage{amsmath,amssymb,amsfonts}%
\usepackage{amsthm}%
\usepackage{mathrsfs}%
\usepackage[title]{appendix}%
\usepackage{xcolor}%
\usepackage{textcomp}%
\usepackage{manyfoot}%
\usepackage{booktabs}%
\usepackage{algorithm}%
\usepackage{algorithmicx}%
\usepackage{algpseudocode}%
\usepackage{listings}%
\graphicspath{ {./} }
\usepackage{subfigure}

\theoremstyle{thmstyleone}%
\theoremstyle{thmstyletwo}%

\theoremstyle{thmstylethree}%

\begin{document}

\title[Article Title]{MAVR-Net: Robust Multi-View Learning for MAV Action Recognition with Cross-View Attention}


\author[1]{\fnm{Nengbo} \sur{Zhang}}\email{zhangnb@student.usm.my}

\author*[1]{\fnm{Hann Woei} \sur{Ho}}\email{aehannwoei@usm.my}

\affil[1]{\orgdiv{School of Aerospace Engineering}, \orgname{Universiti Sains Malaysia}, \orgaddress{ \city{14300 Nibong Tebal}, \state{Pulau Pinang}, \country{Malaysia}}}

\abstract{Recognizing the motion of Micro Aerial Vehicles (MAVs) is crucial for enabling cooperative perception and control in autonomous aerial swarms. Yet, vision-based recognition models relying only on RGB data often fail to capture the complex spatial–temporal characteristics of MAV motion, which limits their ability to distinguish different actions. To overcome this problem, this paper presents MAVR-Net, a multi-view learning-based MAV action recognition framework. Unlike traditional single-view methods, the proposed approach combines three complementary types of data, including raw RGB frames, optical flow, and segmentation masks, to improve the robustness and accuracy of MAV motion recognition. Specifically, ResNet-based encoders are used to extract discriminative features from each view, and a multi-scale feature pyramid is adopted to preserve the spatiotemporal details of MAV motion patterns. To enhance the interaction between different views, a cross-view attention module is introduced to model the dependencies among various modalities and feature scales. In addition, a multi-view alignment loss is designed to ensure semantic consistency and strengthen cross-view feature representations. Experimental results on benchmark MAV action datasets show that our method clearly outperforms existing approaches, achieving $97.8\%$, $96.5\%$, and $92.8\%$ accuracy on the Short MAV, Medium MAV, and Long MAV datasets, respectively.}



\maketitle

\section{Introduction}\label{sec1}

The rapid proliferation of autonomous Micro Aerial Vehicles (MAVs) across domains, such as surveillance \cite{hashim2024advances}, disaster response \cite{javed2024state}, logistics, IoT sensing \cite{zhang2022wi}, and various intelligent application \cite{shekh2025review,gurve2025robot}, has catalyzed the evolution of intelligent and collaborative multi-agent aerial systems. In adversarial or GPS-denied environments \cite{hu2025vision}, motion-based implicit communication \cite{nengbo2025mocomm}, wherein MAVs encode and transmit information through interpretable behaviors, emerges as a transformative paradigm for secure and covert swarm coordination. This approach not only mitigates vulnerabilities to signal jamming and cyber-attacks but also enables real-time, decentralized decision-making. This thereby enhances the resilience and autonomy of MAV swarms in complex missions. However, accurate MAV action recognition, the cornerstone of this communication framework, confronts formidable challenges: the dynamic and intricate nature of MAV behaviors, visual perturbations in complex environments, the need for fine-grained semantic interpretation, and stringent real-time computational constraints in embedded systems. Addressing these challenges represents a pivotal research endeavor with profound academic and practical implications for the development of secure and efficient next-generation aerial systems. 

To achieve robust MAV action recognition, the choice of sensor is paramount. While event cameras, LiDAR, and radar offer advantages in specific scenarios, such as three-dimensional mapping, their high cost, and complex data processing requirements render them impractical for resource-constrained MAV platforms \cite{grandview2023}. In contrast, RGB cameras provide a cost-effective, lightweight, and high-resolution solution for capturing rich visual information, ideally suited for real-time embedded systems. Moreover, RGB video data is readily compatible with established computer vision techniques, offering versatility and scalability for motion recognition tasks. Consequently, this study focuses on leveraging RGB cameras to extract spatio-temporal features from visual data, enabling efficient and economical MAV action recognition.

\begin{figure*}
		\centering 
		\includegraphics[width=0.99\textwidth]{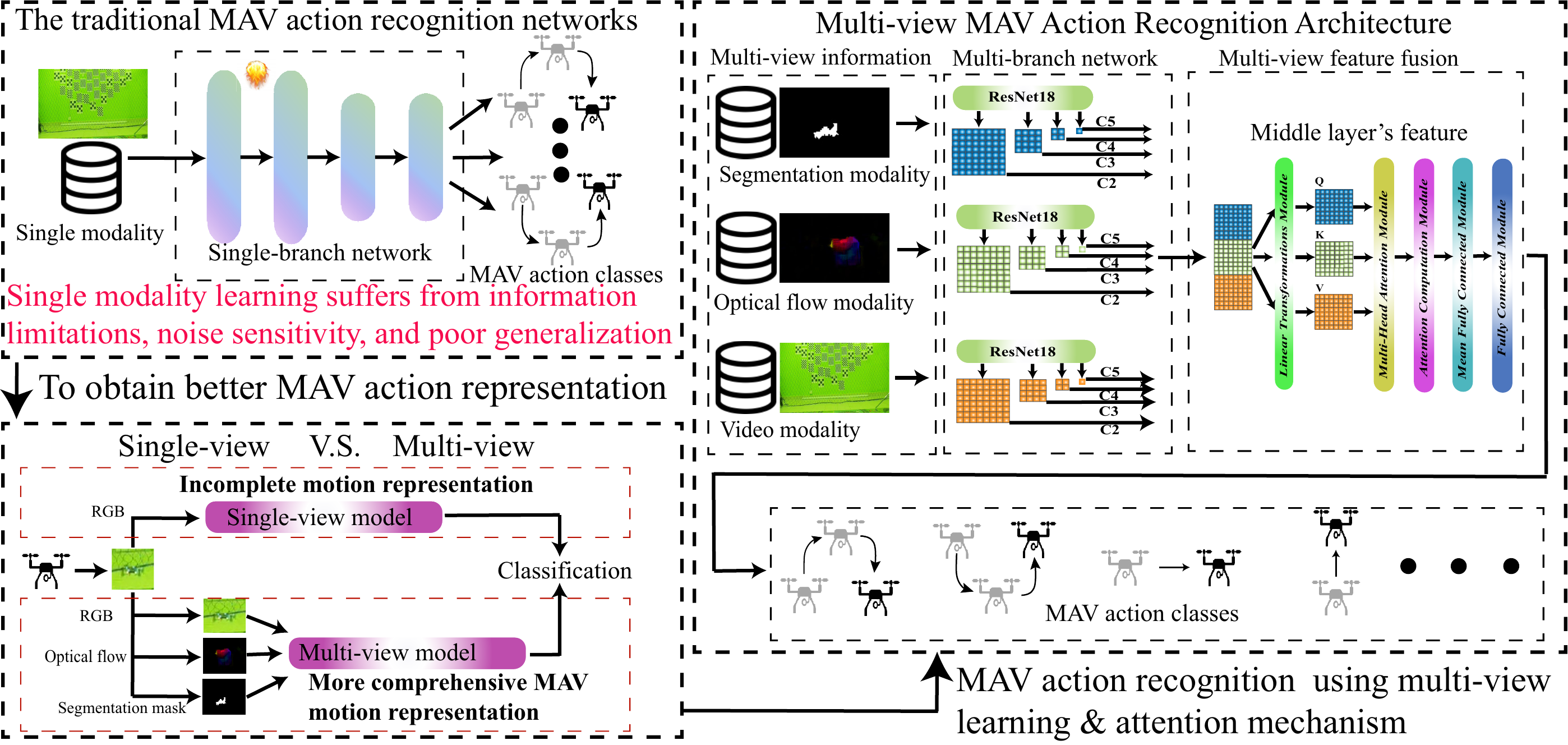}
            
        \caption{Overview of the proposed MAVR-Net architecture compared with traditional single-view models. (Left) Single-branch networks process only RGB frames, making them susceptible to noise and motion ambiguity. (Right) The proposed multi-view learning framework extracts complementary features from RGB, optical flow, and segmentation modalities using parallel branches, followed by feature fusion and cross-view attention to enhance motion discrimination and robustness across diverse flight conditions.}
        \label{fig:firstImage}
\end{figure*}

Given these considerations, recent advances in video-based action recognition provide a strong methodological foundation for MAV action analysis. Visual information processing based on RGB cameras has made remarkable strides in video classification. Seminal approaches, such as Two-Stream Networks \cite{feichtenhofer2016convolutional}, integrate RGB frames and optical flow to capture spatio-temporal features, while 3D Convolutional Neural Networks (3D CNNs) model \cite{tran2015learning} dynamic video evolution through convolutional operations. More recently, Transformer-based models, such as Svformer \cite{xing2023svformer}, have advanced sematic understanding of long-sequence videos. These methods have demonstrated exceptional performance in human action recognition tasks, such as those on the UCF101 and Kinetics datasets, addressing challenges ranging from daily activities to complex behaviors. However, specialized models tailored for MAV action recognition remain scarce. The rapid motion, irregular trajectories, and environmental perturbations (e.g., illumination variations and background clutter) characteristic of MAVs pose unique demands that existing video classification algorithms struggle to meet, underscoring the urgent need for MAV-specific solutions. 

In contrast to human action recognition, MAV motion recognition presents distinct and formidable challenges. First, MAV motion patterns are highly dynamic and diverse, encompassing behaviors such as hovering, sharp turns, and formation transitions, which differ markedly from the structured motions of humans (e.g., walking or waving). Second, MAVs often operate in complex environments where background noise, illumination variations, and occlusions degrade the quality of RGB data, complicating stable feature extraction. Third, the semantic interpretation of MAV actions often hinges on subtle cues, such as rotor speeds or body tilts, demanding heightened motion sensitivity and fine-grained feature resolution beyond the capabilities of conventional models. These challenges collectively delineate a critical research gap in developing robust, MAV-specific action recognition models.  

Multi-view learning  \cite{yan2021deep}, by integrating diverse visual cues offers a powerful framework to address the challenges of MAV action recognition. To analyze MAV's action, RGB frames provide spatial appearance, optical flow captures temporal dynamics, and segmentation masks highlight semantically relevant regions (e.g., rotors or airframes), collectively enabling a comprehensive representation of complex MAV behaviors. Unlike single-modality approaches, multi-view learning enhances robustness to environmental variations and sensitivity to fine-grained actions through cross-view feature fusion and semantic alignment. Its efficacy has been demonstrated across domains: in human action recognition, multi-view frameworks \cite{wu2021mvfnet} combining RGB and optical flow have significantly improved classification accuracy on datasets (e.g., HMDB51); in autonomous driving \cite{fadadu2022multi}, integrating RGB and depth data has enhanced dynamic obstacle detection; and in robotic manipulation \cite{seo2023multi}, multi-view fusion of visual and tactile cues has improved task precision. These successes underscore the potential of multi-view learning to tackle the dynamic, intricate, and environmentally sensitive nature of MAV action recognition, leveraging RGB camera data to achieve high-precision and robust performance.

\begin{figure}
		\centering 
		\includegraphics[width=0.90\textwidth]{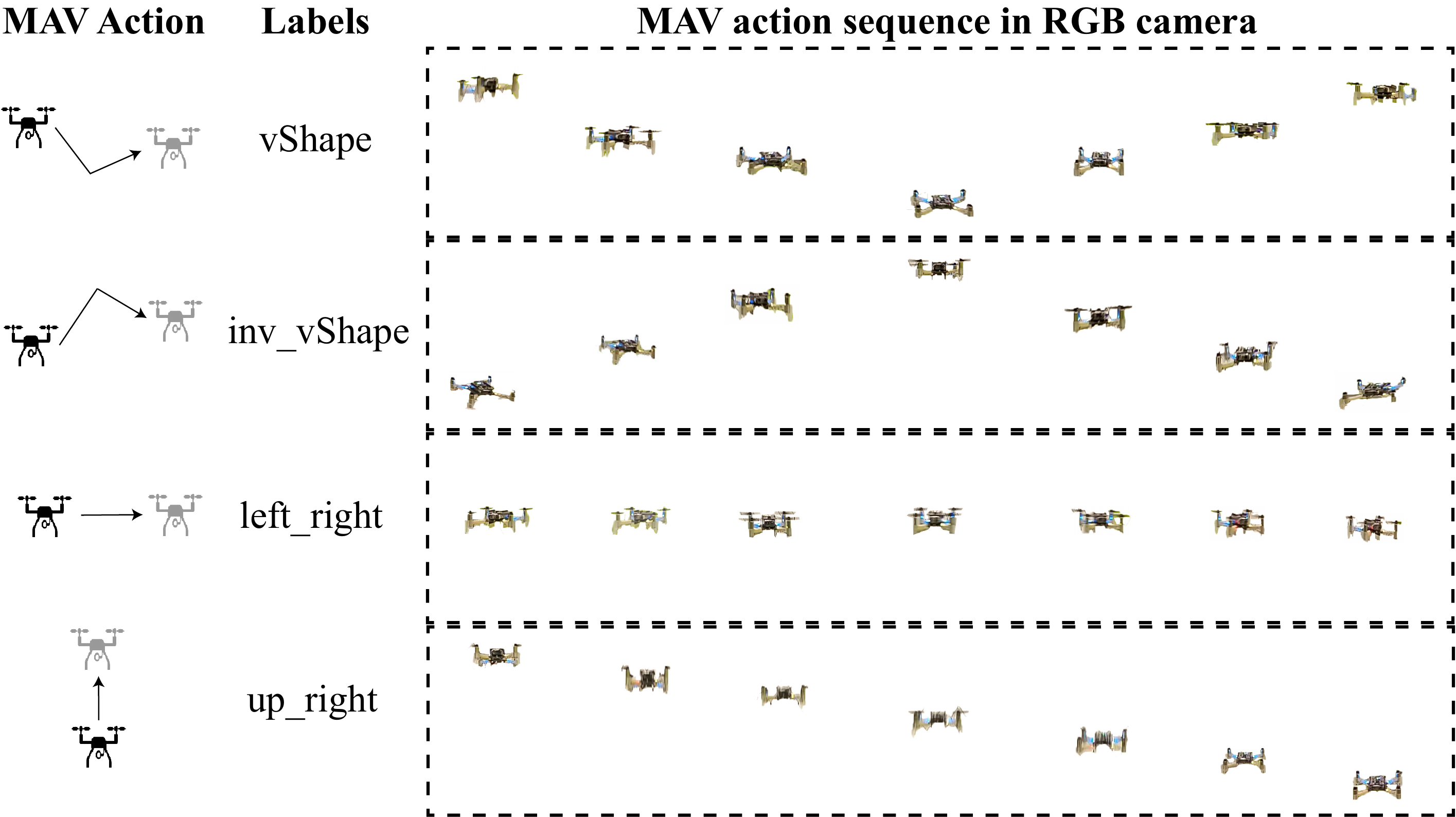}
            
        \caption{Representative motion trajectories of the four MAV action classes: ``vShape'', ``inv\_vShape'', ``left\_right'', and ``up\_right''. These motion types correspond to distinct flight trajectories recorded by the RGB camera and serve as the labeled action categories in the MAVR-Net dataset for training and evaluation.}
        \label{fig:actionSequence}
\end{figure}

Building on the promise of multi-view learning, we propose a novel multi-view neural network framework for MAV action recognition, designed to fully exploit the synergy of RGB frames, optical flow, and segmentation masks. The overview of the proposed algorithm is shown in Fig. \ref{fig:firstImage}. Specifically, our approach employs a unified encoder with residual branches to process each modality, coupled with a multi-scale feature pyramid to capture hierarchical semantic representations. A cross-view attention mechanism dynamically prioritizes feature importance across modalities and scales, while a multi-view alignment loss ensures semantic consistency among heterogeneous views. All experiments are conducted on the collective datasets comprising four types of MAV actions, as shown in Fig. \ref{fig:actionSequence}. This work establishes a robust foundation for secure MAV swarm communication and offers insights for multi-view perception in broader multi-agent systems. Our contributions can be summarized as follows:
\begin{itemize}
  \item A unified multi-view framework (MAVR-Net) is proposed for MAV action recognition, combining RGB, optical flow, and segmentation cues. Compared to traditional single-view learning, MAVR-Net can obtain more comprehensive MAV motion representation.
  \item To address the representation problem in MAV action learning, we introduce a multi-view feature pyramid structure and cross-view attention mechanism for enhanced multi-scale, multi-view information fusion.
  \item To maintain view consistency, we design a novel multi-view alignment loss to encourage semantic coherence across heterogeneous views in MAV action recognition. 
\end{itemize}

The remaining sections of the paper are organized as follows: Section 2 provides a comprehensive overview of MAV communication techniques, vision-based action recognition, and multi-view learning, highlighting the significance of MAV action recognition as a promising research domain. Section 3 details the implementation principles and technical specifics of the proposed network. Experimental results and model analysis are discussed in Section 4. Finally, Section 5 summarizes the key findings of the study and outlines future applications of the proposed method.

\section{Related works}
This section primarily introduces the development of MAV communication techniques, recent works on video-based action recognition methods, and key contributions of multi-view learning in video action recognition.

\subsection{Recent Development of MAV Swarm Communication Techniques}

MAV communication techniques are pivotal for effective control of MAV swarms, enabling enhanced efficiency and precision in swarm operations \cite{aranda2015formation}. Early study \cite{montijano2013distributed} on MAV swarm control strategies proposed distributed approaches for multi-robot data association, allowing independent agents to achieve mutual perception in MAVs. This laid the foundation for robust MAV swarm communication. Additionally, a work \cite{talamali2021less} introduced a local perception-based communication strategy for swarm robots, demonstrating that restricting communication links to local ranges significantly improves control efficiency compared to long-range interactions.
To optimize communication link quality in MAV swarms, signal attenuation and noise issues in the link were addressed in a study \cite{gao2022coverage} through the proposal of a communication method based on pure Nash equilibrium, achieving high-coverage swarm control. Furthermore, a recent work \cite{calvo2024networked} presented a mobile wireless infrastructure with an on-demand allocation design, dynamically adjusting communication routes to ensure high-quality service in MAV swarm communication. MAV swarms employ various communication modes, including point-to-point communication between two MAVs or between a MAV and a base station \cite{liang2019reconfigurable}, ground-to-air communication with ground control platforms \cite{khawaja2019survey}, and GPS-based communication with satellites \cite{hashim2021gps}. Regarding communication efficiency, a work \cite{kim2024codebook} introduced a novel codebook for multi-input multi-output (MIMO) systems tailored for drone communication. This approach enables high-quality, low-complexity data transmission, significantly enhancing the data transfer efficiency of MAV swarms. 

Although traditional MAV communication techniques play a critical role in MAV swarms, they exhibit notable limitations. Firstly, conventional MAV swarm communication relies on radio waves as the transmission medium, which is susceptible to noise interference. Secondly, radio-based MAV communication techniques are vulnerable to eavesdropping by malicious entities. Consequently, several studies have explored alternative MAV communication approaches. For instance, one study \cite{yang2019power} investigated Visible Light Communication (VLC) to simultaneously enable MAV communication and provide illumination in operational environments. Another study \cite{schelle2019visual} leveraged vision-based information to facilitate communication and interaction between MAVs and operators. These efforts have inspired researchers to explore efficient and secure MAV communication techniques.

\subsection{Video-based Action Recognition Methods}
Video action recognition is a fundamental task in computer vision with wide applications in action classification and localization. Traditional video action recognition methods have largely focused on human-centric motion captured from fixed cameras, which poses challenges when directly applied to MAV action recognition scenarios due to viewpoint dynamics, background motion, and MAV motion ambiguity. 

Early work in video recognition introduced 3D convolutional neural networks (3D CNNs) to jointly model spatial and temporal information in videos. For instance, inspired by the success of 2D CNNs in image classification, a study  \cite{tran2015learning} proposed 3D convolutions to directly extract spatio-temporal features from video clips. This approach paved the way for end-to-end action recognition pipelines \cite{huang2023review}. However, 3D CNNs are known for their high computational costs and training complexity due to large parameter sizes. To address these issues, several variants have been proposed. For instance, R-C3D \cite{xu2017r} generates candidate temporal action regions and classifies them by sharing convolutional  features between the proposal and classification pipelines. This allows the action recognition model to have an extremely low parameter count. Besides, 
T-C3D  \cite{liu2018t} further reduces complexity by integrating residual 3D CNNs with hierarchical temporal encoding, capturing multi-granularity features across time domain. As for other action recognition methods, such as a study  \cite{li2019collaborative}, which propose a parameter-efficient spatio-temporal encoding approach by combining 2D CNNs with shared layers to extract complementary video features.

In recent years, attention-based architectures have emerged as strong alternatives to 3D CNNs for video action recognition. For example, transformer-based models such as attention model \cite{gberta_2021_ICML} and SVFormer \cite{xing2023svformer} demonstrate the capability to capture long-range temporal dependencies and contextual relationships in action video sequences. Despite the rapid progress in video action recognition, these methods are not directly suitable for MAV-based action understanding. Unlike human action datasets, MAV action data involves flying agents, where motion cues are more subtle and affected by ego-motion, scale variation, and inter-agent synchronization. 

\subsection{Multi-View Representation Learning in Video Action Recognition}
To improve robustness and generalization in video action recognition, multi-view has emerged as a powerful strategy. Multi-view learning, compared to traditional single-view learning, leverages the complementarity of information across different views to effectively overcome the information bottleneck of a single view, thereby achieving more robust video action recognition. A classic two-stream networks \cite{simonyan2014two} process RGB frames and optical flow separately to capture appearance and motion cues. It achieves the best video action recognition results of that year. Subsequent research, such as SlowFast \cite{feichtenhofer2019slowfast}, advanced this paradigm by fusing view information with varying temporal resolutions, enabling the model to effectively capture short-term and long-term dynamic features of video actions. These approaches highlight the benefits of multi-view information in action recognition problem. Building upon this foundation, researchers have explored cross-view attention and feature fusion strategies to improve interaction between different views. For instance, a study \cite{sun2019learning} introduced modality-specific encoders with attention-based fusion for video-language alignment, while another work \cite{kim2020modality} proposed shifting attention network for multi-view video question answering. These methods demonstrate the effectiveness of learning coordinated and complementary representations from distinct modalities or views.

In parallel, multi-view alignment and contrastive learning has gained attention for its ability to reduce view-level semantic gaps. Inspired by contrastive representation learning, models like CLIP \cite{radford2021learning} and VATT \cite{akbari2021vatt} jointly align features across vision and language (or different visual views) using instance-level or set-level contrastive losses. These methods effectively enforce semantic consistency among views and enhance feature discriminability. In contrast to existing works mainly focused on information from different modalities, our method targets multi-view MAV action recognition, where aligning fine-grained inter-view features is critical due to subtle motion patterns and dynamic perspectives. 


\section{The MAVR-Net Framework}

This section describes the basic multi-view learning theory and presents the proposed multi-view MAV action recognition framework, i.e., MAVR-Net, including the overall architecture, multi-View feature extraction, multi-view feature pyramid fusion module, cross-view attention mechanism, multi-view alignment loss, and the final training objective of our proposed network.

\subsection{The Basic Principle of Multi-View Learning}
Multi-view learning is a powerful paradigm that leverages multiply complementary representations of the same instance to improve recognition performance and enhance model generalization. Each view may correspond to a different modality, abstraction, or perspective; For example, in MAV motion video, RGB frames can capture appearance cues of MAV. Besides, optical flow features describe MAV's motion dynamics and direction. Then, segmentation masks information can emphasize MAV's spatial structures in the sky. By jointly modeling these diverse views, multi-view learning enables the recognition network to build a more complete and discriminative understanding of MAV actions.

The specific principle of multi-view learning can be derived from the following mathematical description. Formally, suppose there are $M$ views, and the $m-th$ view is represented as an input vector  $x_m \in \mathbb{R}^{d_m}$, where $m = 1, 2, ..., M$. The objective of multi-view representation learning is to map these inputs into a shared latent space as a joint representation $z \in \mathbb{R}^d$ in Eq. \ref{eq:multi_learning}.
\begin{equation}
\label{eq:multi_learning}
\begin{split}
z = f(x_1,x_2,...,x_M; \theta),
\end{split}
\end{equation}
where $f$ is the fusion mapping function, typically implemented using a deep neural networks, and $\theta$ denotes the learnable model parameters. In the applications, the discriminative ability of the multi-view model $f$ typically improves as the number of input views $m$ increase.   

In multi-view learning, to enable effective collaboration among different views, it is often necessary to design appropriate inter-view fusion strategy. Existing multi-view fusion strategies fall into three categories: early fusion, late fusion, and joint representation learning fusion. Early fusion integrates features from all views at the input stage, combining raw or preprocessed data into a unified representation before model training. Late fusion strategy processes each view independently through separate models and aggregates their outputs, such as predictions or decisions, at the final stage. Joint representation learning fusion method learns a shared latent representation across views during training, optimizing the model to capture complementary information from all views simultaneously.


\subsection{Overall Architecture for MAV Action Recognition }
\begin{figure*}
		\centering 
		\includegraphics[width=0.99\textwidth]{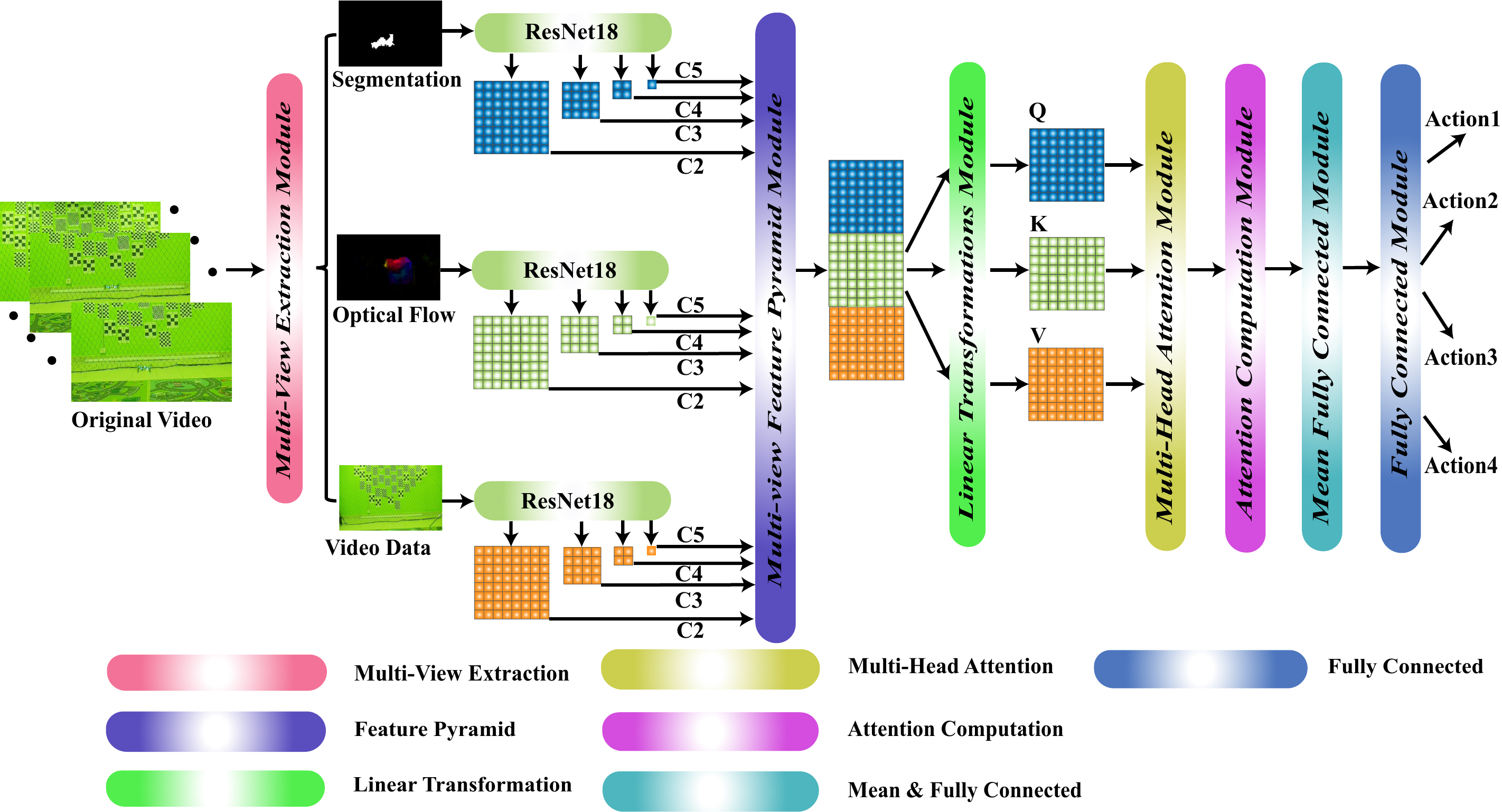}
            
		\caption{The whole pipeline of multi-view learning framework. From left to right, the raw MAV video action data is processed by multi-view extraction, multi-view feature pyramid process and etc. Besides, $Q$, $K$, and $V$ represent the Query, Key, and Value feature matrices used in the multi-head attention mechanism.  } 
        \label{fig:pipeline}
\end{figure*}

To achieve accurate video-based action recognition for MAV, it is critical to efficiently exploit camera sensor data. Conventional action recognition methods predominantly rely on visual data alone, making them vulnerable to noise and motion blur caused by high-speed MAV movements. Moreover, relying solely on a single visual modality creates an information bottleneck, as it fails to provide comprehensive motion information from multiple perspectives. Departing from existing approaches, we propose a multi-view learning strategy that leverages complementary modalities for robust MAV action recognition. Specifically, we introduce a multi-view fusion framework integrating three complementary cues: (1) RGB views to capture visual appearance of MAVs, (2) optical flow to model their motion dynamics, and (3) segmentation masks to delineate MAV shape information, effectively filtering out background interference. Our motivation for employing multi-view learning in MAV action recognition is inspired by the sophisticated perception of bees within a swarm, where individual bees integrate multiple sensory cues—such as visual patterns, motion trajectories, and spatial relationships—to recognize and respond to the behaviors of other bees. This biological insight highlights the inherent limitations of relying on individual modalities in complex MAV scenarios, where RGB views capture appearance but lack motion dynamics, optical flow encodes motion but misses detailed textures, and segmentation masks isolate shapes yet struggle with dynamic backgrounds. By adopting a multi-view framework that integrates these complementary modalities, our approach mirrors the robust perceptual strategy of bees, enabling accurate and reliable MAV action recognition in challenging, dynamic environments. The proposed architectures (Fig.  \ref{fig:pipeline}) facilitates cross-view feature interaction while preserving discriminative spatio-temporal representations. This fusion mechanism is particularly effective under viewpoint variations, scale differences, and cluttered backgrounds. Experimental results also demonstrate that our method significantly improves MAV action recognition accuracy by synergizing multi-view information.   

As illustrated in Fig. \ref{fig:pipeline}, our MAV action recognition pipeline extracts three view-specific modalities from the original video data, including RGB, optical flow, and segmentation masks. Specifically, the RGB data stream is obtained through identity mapping, while the MAV optical flow view data is obtained through dense optical flow calculation \cite{OpenCV2025}. Then, the MAV mask information is obtained through moving object segmentation \cite{yao2020video}. Each modality is processed by an independent ResNet-18 branch with a 3D convolutional encoder to extract spatio-temporal features (denoted as C2 to C5). These modality-specific features are then fed into a shared Feature Pyramid Fusion Module (FPM) to generate multi-scale MAV action representations. The fused pyramid features are concatenated across all views, followed by a cross-attention mechanism to model inter-view dependencies and highlight strongly correlated features. Finally, the discriminative features are aggregated via global average pooling and processed by a fully connected layer to predict the target MAV action labels.

\subsection{Multi-View Feature Extraction and Feature Pyramid Fusion Module}
To extract complementary multi-view modalities from the original MAV action video, we employ established computer vision algorithms to derive three synchronized representations: RGB frames, optical flow fields, and segmentation masks. While spatio-temporally aligned, these modalities provide distinct visual cues from the homogeneous RGB source. For example, the RGB frames preserve appearance details of MAV through identity transformation, the dense optical flow fields (computed via Farneback's algorithm) explicitly encode MAV motion dynamics, and the segmentation masks (obtained through moving object segmentation) isolate the MAV while suppressing background interference. Each modality is independently processed by a 3D ResNet-18 backbone, where 3D convolutions model spatio-temporal patterns over 16-frame clips, enabling explicit separation of appearance feature (RGB), motion patterns (optical flow), and structural saliency (masks) to construct a rich, complementary feature space for robust MAV action recognition.    

In MAV action recognition, as for visual sensors, the varying observation altitudes and distances cause targets and actions to appear at different scales within the same video sequence. This makes single-scale feature extraction methods ineffective in simultaneously capturing fine-grained motion details at long distances (e.g., MAV actions from afar) and coarse-grained motion patterns at close ranges (e.g., MAV movements up close). Such limitations significantly degrade the robustness of MAV action recognition models in real-world scenarios. To address this problem, we introduce a Multi-View Feature Pyramid Module (MVFPM), which constructs multi-scale representations of MAV video features under different views. Unlike the feature pyramid research in a study \cite{lin2017feature}, the pyramid features constructed in this study are specifically designed for multi-view features of MAV video data. The complexity and richness of the features processed significantly surpass those of traditional feature pyramid models.

\begin{figure}[!t]
		\centering 
		\includegraphics[width=0.7\textwidth]{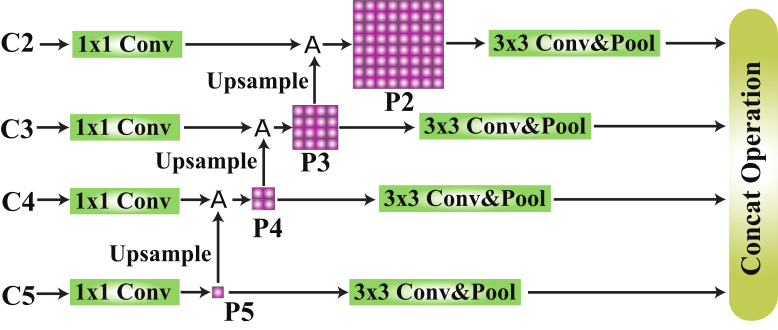}
            
		\caption{A detailed overview of the feature pyramid. C2 to C5 represent multi-stage features from the view encoding network (ResNet18). ``Upsample'' denotes the upsamling operation, while ``A'' represents the addition operation. } \label{fig:FPN}
\end{figure}

This simple yet effective module ensures accurate action recognition across varying MAV scales. As illustrated in Fig. \ref{fig:FPN}, the MVFPM leverages hierarchical feature maps (C2-C5) extracted from the backbone network (ResNet18) to generate a set of feature maps (P2-P5) with consistent resolution but distinct semantic and spatial details. Through top-down aggregation and lateral connections, each level of the pyramid integrates high-resolution spatial information with rich semantic cues, enhancing scale-invariant feature learning. The architecture details of MVFPM is following. As for input layer (C2-C5), Multi-stage spatio-temporal feature maps from ResNet18, where C2 has high resolution but low semantics, while C5 exhibits low resolution but deep semantics. Then, each input undergoes 1x1 convolution for channel reduction. C3-C5 are upsampled to match the resolution of the preceding stage. The upsampled features are fused with the corresponding lower-level features via element-wise addition (e.g., upsampled C5 merges with C4). A 3x3 convolution refines the fused features, followed by pooling to produce the final pyramid levels (P2-P5). Finally, multi-scale feature maps where P2 retains the finest spatial details, and P5 encodes the highest-level semantics, collectively capturing scale-specific discriminative patterns from MAV motion. 

To leverage multi-view inputs, three views (segmentation masks, optical flow, and raw video frames) are independently processed, with their respective C2-C5 features extracted using ResNet18. The multi-scale features from each modality are then concatenated, forming a unified multi-view feature pyramid. This representation not only encapsulates complementary MAV motion cues from different modalities but also adapts to scale variations via the MVFPM. Compared to existing action recognition models, our multi-view MAV action recognition framework with MVFPM demonstrates superior robustness to scale changes, attributed to the proposed multi-view feature pyramid design.

\subsection{Cross-View Attention Mechanism for Multi-view MAV Action Features}

To enable multi-view networks to effectively learn contextual consistency among features from different views under MAV action during flight, a cross-view attention model tailored for MAV action recognition is proposed in this paper. This model specifically processes the fused multi-view feature maps output by the MVFPM. The primary motivation for designing this module is that MAV actions may exhibit distinct characteristics in certain views while appearing ambiguous in others. For instance, motion blur often occurs in video views, whereas optical flow views typically provide stronger discriminative features. Consequently, efficiently integrating these heterogeneous data streams and learning comprehensive MAV motion information from fused multi-view feature maps pose significant challenges for robust MAV action recognition. To address these issues, we introduce a Transformer-inspired Cross-View Attention Mechanism (CVAM) that dynamically integrates multi-view features, enhancing the model's ability to recognize actions in MAV video.

The proposed Cross-View Attention Mechanism (CVAM) leverages the self-attention mechanism of Transformers to model contextual relationships of MAV action among different views. Specifically, after extracting multi-scale features in the form of a feature pyramid for each view (RGB, optical flow, and mask) using a ResNet-18 backbone network, these features across different scales are concatenated. Following the extraction of multi-scale information from different views, a multi-head attention modules is employed to fuse these features, denoted as $F_{mv}$. During the fusion process, the fused features are duplicated to serve as the query, while two additional copies are used as the key and value, respectively. This process is formalized in Eq. \ref{eq:cvam}.

\begin{equation}
\begin{split}
F_{mv}(Q, K, V) = \text{softmax}\left(\frac{QK^T}{\sqrt{d}}\right) \cdot V,
\end{split}
\label{eq:cvam}
\end{equation}
where $Q$,$K$, and $V$ denote query, key, and value projections of the feature vectors, and $d$ is the dimensionality of the key vectors to stabilize gradient scaling. In implementation, for a batch size $B$ with $T$ frames, the MVFPN outputs are reshaped into a tensor of shape $(B, T, 3072)$, where 3072 represents the concatenated feature dimension across the three modalities (1024 per modality). A multi-head attention module with 12 heads processes this tensor to produce a context-aware representation capturing cross-view interactions. The attention output is averaged over the temporal dimension and passes through two fully connected layers to predict the MAV action class.

The CVAM dynamically integrates these FPM's features, enabling the model to prioritize relevant MAV action cues. Thus, the CVAM significantly improves MAV action recognition by enabling adaptive feature fusion. Unlike traditional concat fusion method, CVAM dynamically adjusts feature weights based on the input, ensuring that critical cues, such as motion contours in optical flow or salient regions in masks, are prioritized. This is particularly crucial in MAV scenarios, where varying altitudes or occlusions may obscure certain modalities. By modeling inter-view dependencies, CVAM enhances the model’s robustness to such challenges. Additionally, the multi-head attention mechanism captures diverse interaction patterns, further enriching MAV action's feature representations.

\subsection{Multi-View Alignment Loss and Training Objective}
In training the proposed action recognition network, effectively integrating complementary views (RGB, optical flow, and segmentation masks) is critical for achieving robust MAV action recognition. However, due to the distinct visual cues (e.g., appearance for RGB view, motion for optical flow view, or structural information for segmentation view) across these views, differences in their feature representation can impede training convergence, hindering effective multi-view information fusion. To address this challenge, a multi-view alignment loss function is introduced for MAV action recognition. Its primary goal is to enhance cross-view semantic consistency, ensuring that the feature representation of RGB, optical flow, and masks are aligned in a shared latent space, thereby promoting cross-view synergy. This results in high accuracy MAV action recognition. The detailed computation of this loss function is describe in the next.  

Given modality-specific features $f^{RGB}$,$f^{Flow}$, and $f^{Mask}$ extracted from previous module, we compute their mean-pooled representations over the temporal dimension and normalize them to unit length. The alignment loss leverages a contrastive learning objective, where pairwise similarities between modalities are computed as $\text{sim}(f_i, f_j) = f_i \cdot f_j^T / \tau$, with $\tau=0.07$ as the temperature parameter. For a batch of size $B$, we calculate similarity matrices for each pair (RGB-flow, RGB-mask, flow-mask) and apply cross-entropy loss against ground-truth labels (diagonal indices $[0, ..., B-1]$), encouraging positive pairs (same-sample features across views) to align while pushing negative pairs apart. The total loss is in Eq. \ref{eq:align}.
\begin{equation}
\begin{split}
\mathcal{L}_{align} = \frac{1}{3} \Big[ &\mathcal{L}_{CE}(\text{sim}(f^{RGB}, f^{Flow}), y) \\
& +  \mathcal{L}_{CE}(\text{sim}(f^{RGB}, f^{Mask}), y)  \\
& + \mathcal{L}_{CE}(\text{sim}(f^{Flow}, f^{Mask}), y) \Big],
\end{split}
\label{eq:align}
\end{equation}
where $\mathcal{L}_{CE}$ is the cross-entropy loss and $y$ denotes the identity labels. This loss enhances our framework by aligning modality-specific features before fusion in the MVFPM and CVAM, ensuring coherent multi-scale representations. By reducing representational inconsistencies, it improves the model’s robustness to MAV-specific challenges like viewpoint variations and occlusions, leading to more accurate action recognition.

Although the cross-view attention mechanism integrates features from RGB, optical flow, and mask modalities, the attention mechanism may overfocus on specific features or views, leading to overfitting or collapse, which reduces generalization of MAV action recognition. The attention regularization loss $\mathcal{L}_{att}$ maximizes the entropy of attention weights to promote a uniform distribution, enhancing robust feature integration and improving MAV action recognition accuracy under challenges like scale variations and occlusions. Eq. \ref{eq:l_att} provides detailed calculations. Attention weight $a_{ij}$ represents the strength of association between the \(i\)-th source view element and the \(j\)-th target view element. $N$ is the number of elements in the source view, while $M$ represents number of the elements in the target view. $\mathcal{L}_{att}$ minimizes negative entropy to encourage a more uniform attention weight distribution, preventing over-reliance on each view. 

\begin{equation}
\begin{split}
\mathcal{L}_{att} = \frac{1}{N} \sum_{i=1}^N \sum_{j=1}^M a_{ij} \log a_{ij}.
\end{split}
\label{eq:l_att}
\end{equation}

For the final training objectives of multi-view MAV action recognition, this paper introduces a cross-attention regularization loss. It enables cross-view feature alignment and balanced attention distribution across different views during the training stage. The total loss function of the proposed model is described in Eq. \ref{eq:total}. 
\begin{equation}
\begin{split}
\mathcal{L}_{total} = \mathcal{L}_{cls} + 
 \lambda_1 \mathcal{L}_{align} + \lambda_2 \mathcal{L}_{att},
\end{split}
\label{eq:total}
\end{equation}
where $\mathcal{L}_{cls}$ is the standard cross-entropy loss for predicting MAV action labels, $\mathcal{L}_{align}$ enforces semantic consistency across RGB, optical flow, and mask modalities (as detailed in Eq.~\ref{eq:align}), and $\mathcal{L}_{att}$ regularizes the cross-view attention mechanism to prevent collapse or overfitting by maximizing entropy of attention weights. The hyper-parameters $\lambda_1$ and $\lambda_2$ control the contribution of each loss term.

\section{The experiments}
This section details several experiments for multi-view MAV network. First, the setup and data collection for the MAV action recognition experiment are detailed. Next, a comprehensive overview of the evaluation metrics is provided. Finally, ablation studies for the proposed module are presented, along with a comparison of its recognition performance against mainstream action recognition methods.

\subsection{Dataset Collection and Experimental Settings}
\begin{figure}
		\centering 
		\includegraphics[width=0.7\textwidth]{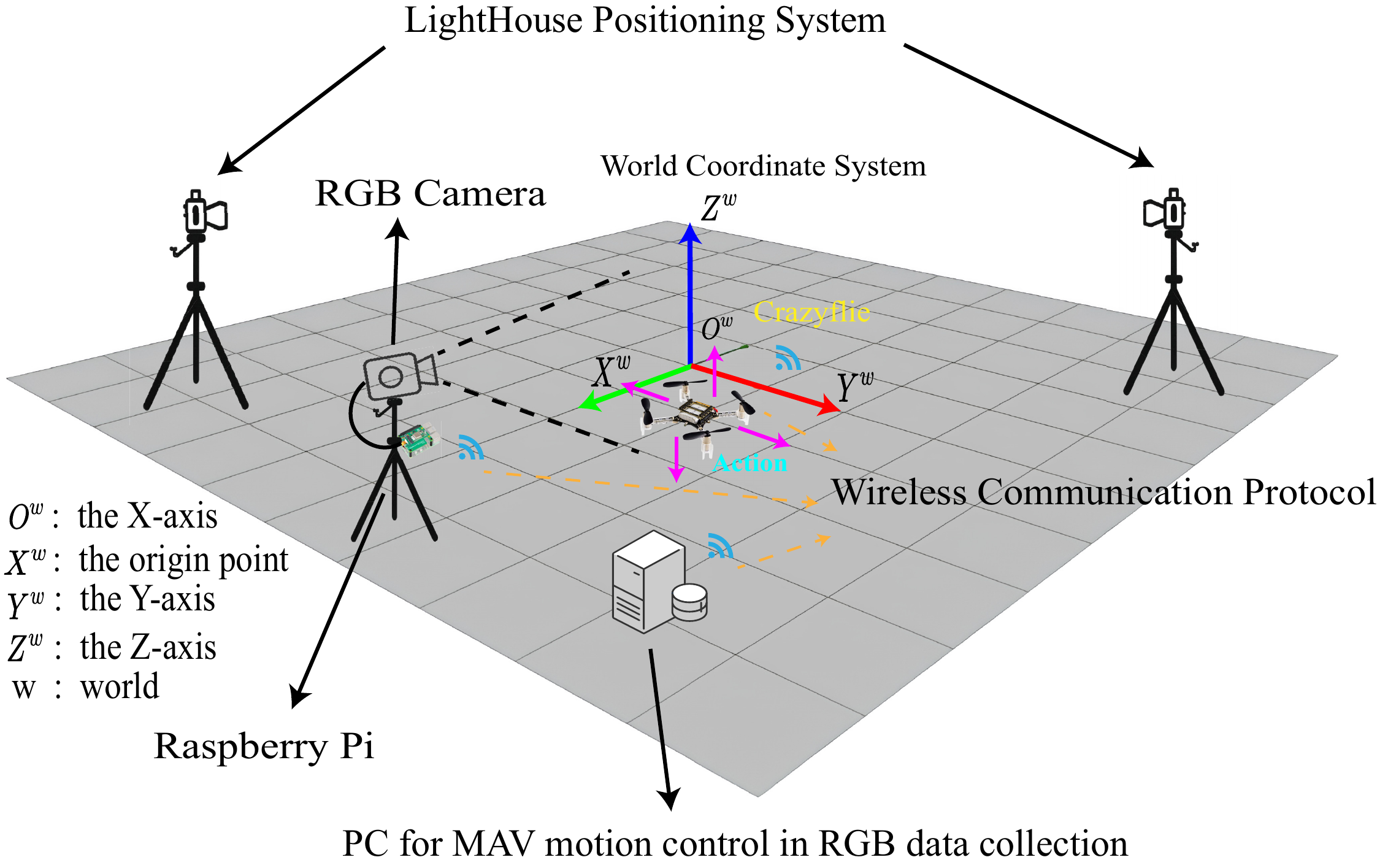}
		\caption{Schematic diagram of the data acquisition system for MAV action dataset.} \label{fig:dataCollect}
\end{figure}

To construct the MAV action recognition dataset\footnote{Dataset and codes: \url{https://github.com/iAerialRobo/MAVR-Net.git}}, we utilized Bitcraze CrazyFlie 2.1+ MAV equipped with Lighthouse positioning decks, achieving sub-millimeter precision control within a 5x5x3 meter indoor obstacle-free environment (see Fig. \ref{fig:dataCollect}) under consistent lighting. The MAV is controlled via a personal computer running Ubuntu 22.04 with the cfclient package \cite{cfclient_repo}, while video footage was captured by a Raspberry Pi 4.0 (running Ubuntu 20.04) equipped with a 640x480 resolution RGB camera (60-degree field of view) at 30 fps. As for MAV action dataset, we collected data on four motion patterns: vertical motion, horizontal motion, V-shaped motion, and inverse V-shaped motion, each with a flight radius of 0.3 meters and a speed of 0.2 m/s. Around 1500 video samples per scale are recorded at three camera-to-MAV distances (Short: 1m, Medium: 1.5m, Long: 2m), each annotated with a single motion label for MAV-to-MAV communication via actions. These samples were used to construct small-scale, medium-scale, and large-scale MAV motion datasets to evaluate the robustness of action recognition algorithms across spatial scales.

As for experimental settings in MAV action recognition, we utilized the PyTorch deep learning framework, designing a model architecture where each view is processed by an independent ResNet18 network for feature extraction. The model is trained for 25 epochs with a learning rate of 0.01 and a batch size of 64, using the SGD optimizer (momentum 0.9, weight decay 0.0001). Then, we apply $\lambda_1$ (0.5) and $\lambda_2$ (0.1) to control the weight of each loss term in Eq. \ref{eq:total}. During the data preparation phase, the MAV action dataset is split into training and testing sets with a ratio of 2:1 (training:testing), resulting in approximately 1000 training samples and 500 testing samples per action category. Video frames were normalized to $[0,1]$, cropped to 224x224 pixels, and augmented with random horizontal flipping. Training was conducted on a computing node equipped with a NVIDIA RTX 4090 (48GB) GPU.

\subsection{Experimental Metrics }
To assess the performance of our multi-class classification model, we use four key metrics: Accuracy, Precision, Recall, and F1-Score, which provide insights into different aspects of model effectiveness for a dataset with $C$ classes and $N$ total samples. Accuracy measures the proportion of correctly classified samples, suitable for balanced datasets but less reliable for imbalanced ones. Precision quantifies the reliability of positive predictions, critical when false positives are costly. Recall evaluates the ability to identify all relevant samples, essential when missing positives is undesirable. The F1-Score, the harmonic mean of Precision and Recall, balances their trade-off and is particularly useful for imbalanced datasets. These metrics are macro-averaged for multi-class tasks and defined as follows in Eq. \ref{eq:metrics}.

\begin{equation}
\begin{split}
\text{Accuracy} &= \frac{\sum_{i=1}^{C} \text{TP}_i}{N}, \\
\text{Precision}_{\text{macro}} &= \frac{1}{C} \sum_{i=1}^{C} \frac{\text{TP}_i}{\text{TP}_i + \text{FP}_i}, \\
\text{Recall}_{\text{macro}} &= \frac{1}{C} \sum_{i=1}^{C} \frac{\text{TP}_i}{\text{TP}_i + \text{FN}_i}, \\
\text{F1}_{\text{macro}} &= \frac{1}{C} \sum_{i=1}^{C} \left( 2 \cdot \frac{\text{Precision}_i \cdot \text{Recall}_i}{\text{Precision}_i + \text{Recall}_i} \right),
\end{split}
\label{eq:metrics}
\end{equation}
where $\text{TP}_i$, $\text{FP}_i$, and $\text{FN}_i$ are the true positives, false positives, and false negatives for class $i$, respectively, with $\text{Precision}_i = \frac{\text{TP}_i}{\text{TP}_i + \text{FP}_i}$ and $\text{Recall}_i = \frac{\text{TP}_i}{\text{TP}_i + \text{FN}_i}$.

\subsection{Comparison of MAV Action Recognition Accuracy with State-of-the-Art Algorithms}

In order to evaluate MAV action recognition performance of the proposed model, we conducted comparative experiments on three MAV action datasets with other state-of-the-art algorithms. The algorithms include C3D \cite{tran2015learning}, CoST \cite{li2019collaborative}, and Svformer \cite{xing2023svformer}. The overall recognition results are tested at three different scales, and the test results are presented in Table \ref{tab:comparsion_other}. From the Table \ref{tab:comparsion_other}, it can be observed that the performance of the algorithms in recognizing MAV actions varies across different distances. In long-distance scenarios, all models exhibit a decline in recognition accuracy accompanied by an increase in the standard deviation of accuracy. However, the proposed model in this study maintains a relatively high level of accuracy despite the existence of scale variations.  

\begin{figure}
		\centering 
		\includegraphics[width=0.6\textwidth]{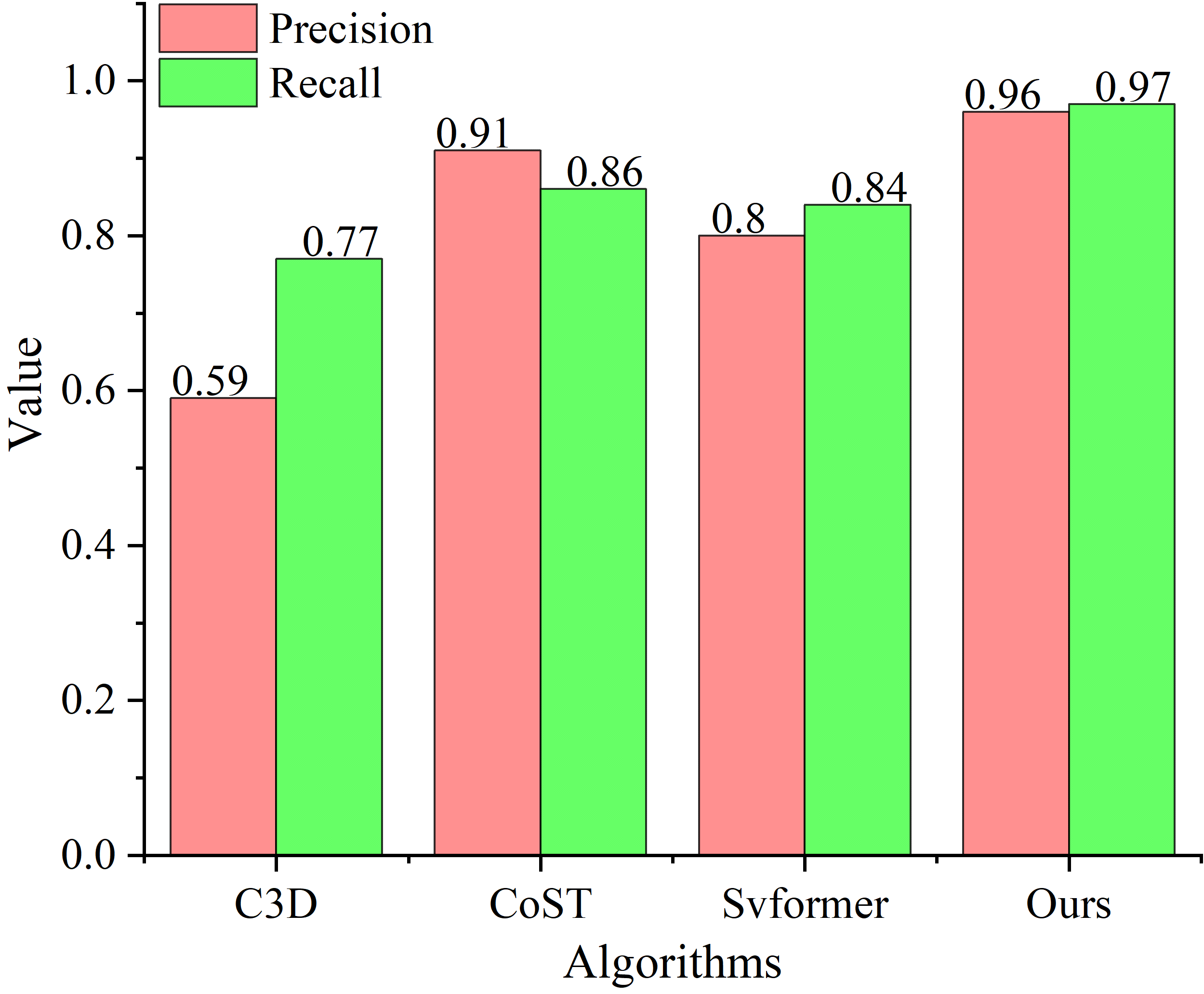}
		\caption{The comparison of precision and recall metrics in the medium MAV action dataset.} 
        \label{fig:pre_recall}
\end{figure}

\begin{table}[h]
    \centering
    \caption{Results of Action Recognition Methods in Comparison.  }
    \label{tab:comparsion_other}
    \begin{tabular}{cccc}
    \hline
    Algorithms & Short & Medium & Long \\ \hline
    C3D \cite{tran2015learning}  &$70.4\%\pm 7.8\%$ &$64.5\%\pm 8.4\%$  & $62.1\%\pm 16.4\%$  \\ 
    CoST \cite{li2019collaborative} &$87.5\%\pm 9.5\%$&$87.1\%\pm 11.1\%$&$84.2\%\pm 13.5\%$\\
    Svformer \cite{xing2023svformer}   &$89.1\%\pm 10.4\%$&$82.2\%\pm 15.1\%$&$70.4\%\pm 19.7\%$    \\
    Ours  &$97.8\%\pm 2.8\%$&$96.5\%\pm 3.9\%$ &$92.8\%\pm 4.7\%$      \\
    \hline
    \end{tabular}
\end{table}

Besides, to assess the effectiveness of our proposed model in MAV action recognition, we conducted experiments on the collected Medium distance MAV action recognition dataset, we adopted Precision and Recall as evaluation metrics, where Precision measures the accuracy of the model in predicting action categories, and Recall assess the model's ability to identify true action instances. To comprehensively compare model performance, we compared three state-of-the-art baseline models (C3D, CoST, and Svformer) with our proposed model (ours). Fig. \ref{fig:pre_recall} illustrates the Precision and Recall performance of different models on the MAV action recognition task. The x-axis represents the models (C3D, CoST, Svformer, and our proposed model), while the y-axis indicates the values of Precision and Recall, ranging from 0 to 1. The red bars represent Precision, and the green bars represent Recall. As shown in the figure, C3D achieves a Precision of 0.59 and a Recall of 0.77, CoST achieves a Precision of 0.91 and a Recall of 0.86, and Svformer achieves a Precision of 0.80 and a Recall of 0.84. It is evident that our proposed model significantly outperforms the baseline models in both Precision and Recall. Compared to C3D algorithm, our model improves prediction accuracy by approximately 40 percentage points while maintaining a high recall rate. Although Cost and Svformer exhibit relatively balanced performance in Precision and Recall, they still fall short of our model. Notably, our model achieves a Precision of 0.96 and a Recall of 0.97, indicating near-perfect accuracy in MAV action prediction. 

\begin{figure}
		\centering 
		\includegraphics[width=0.6\textwidth]{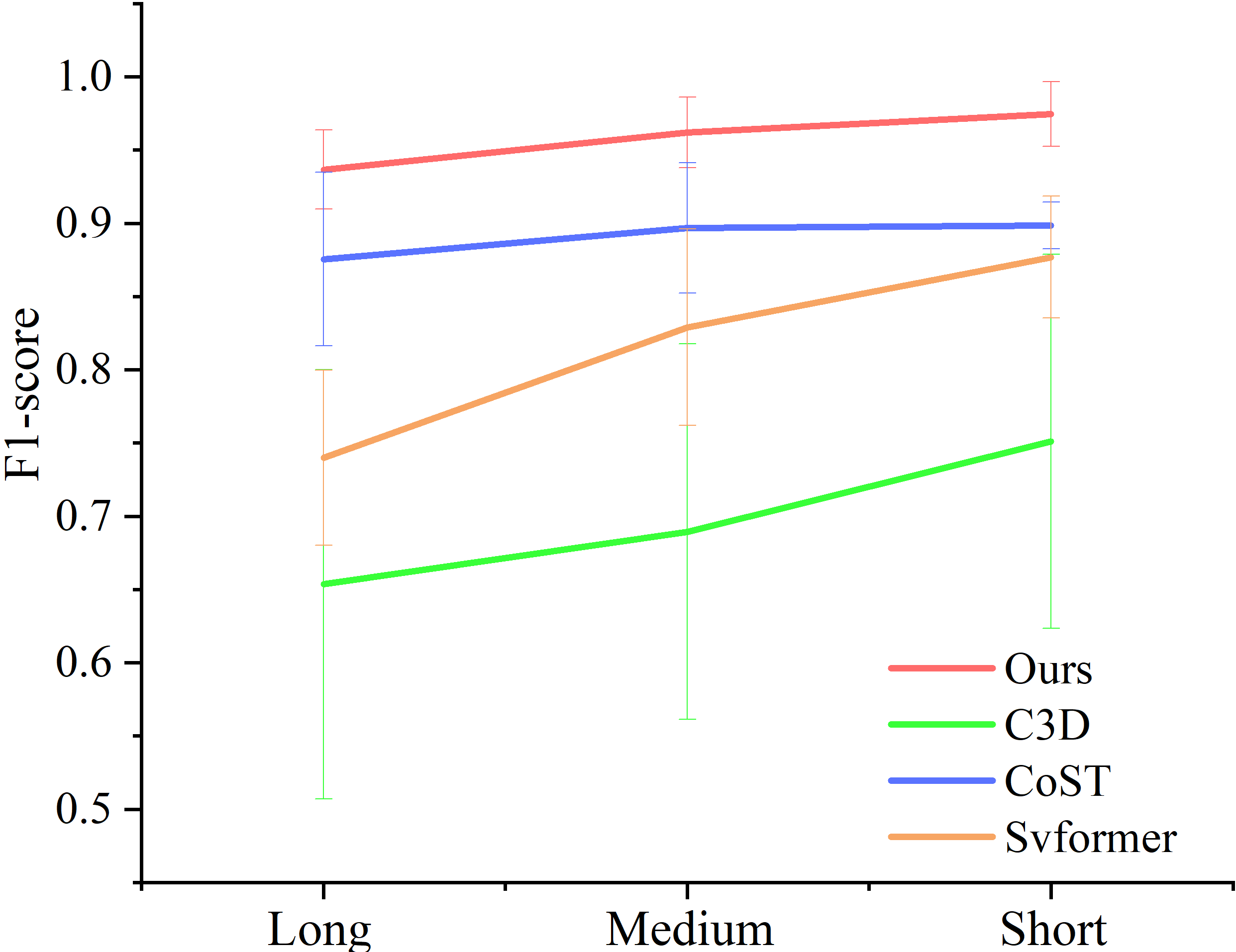}
		\caption{F1-Score comparison across different scales of MAV action recognition datasets.} 
        \label{fig:f1}
\end{figure}

To further assess the overall performance of the models, we introduced the F1-score, which is the harmonic mean of precision and recall, to measure the balance between precision and recall. As shown in Figure \ref{fig:f1}, our model achieves F1-scores close to 0.95 across datasets with different MAV observation distances (Long, Medium, and Short), significantly surpassing the performance of C3D, CoST, and Svformer. Moreover, our model maintains small score fluctuations when recognizing actions at varying MAV scales. This not only highlights its stability in MAV action recognition but also underscores its robustness against changes in MAV action scales.

\subsection{MAV Action Classification Performance and Training Stability}

\begin{figure}
		\centering 
		\includegraphics[width=0.65\textwidth]{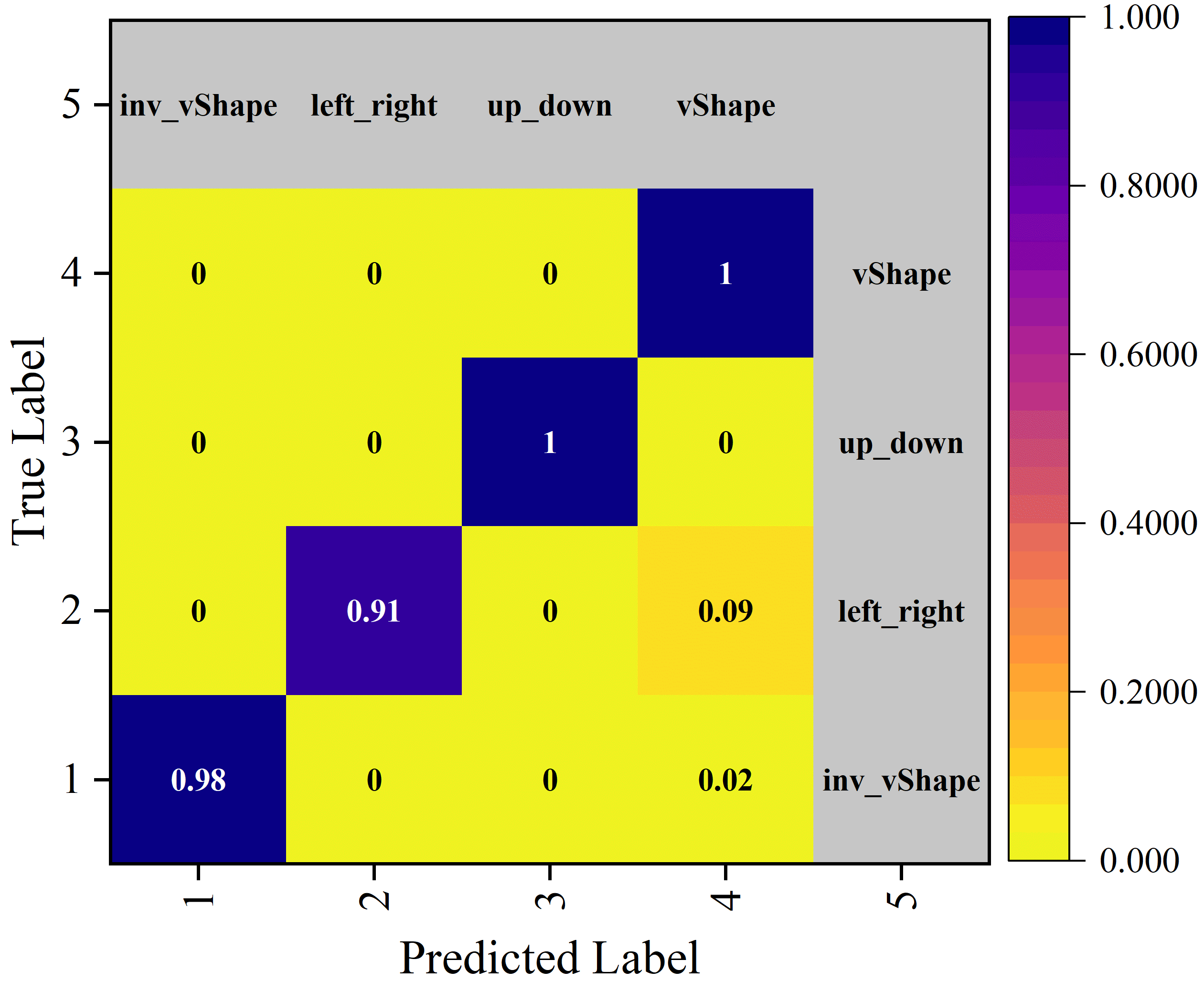}
		\caption{Confusion matrix of MAVR-Net in short distance MAV action dataset.} \label{fig:mavr_confusion}
\end{figure}

To evaluate the proposed model's performance on distinct MAV actions and training stability on the MAV action recognition task, this section presents the confusion matrix visualization for classification results and the convergence curves of accuracy and loss during training stage. In Fig. \ref{fig:mavr_confusion}, we provide the confusion matrix for MAV action recognition in short-range scenarios (Short). The distribution of classification results shows that MAVR-Net effectively distinguishes between the ``vShape'' and ``up\_down'' motion patterns. However, for the ``left\_right'' action label, there is a 0.09 probability of misclassification as ``vShape''. This is due to the similarity between ``left\_right'' and ``vShape'' motions, both involving left-to-right movement, posing a challenge for the MAV action classification model. Similarly, for the ``inv\_vShape'' action label, a small number of samples are misclassified as ``vShape'' because of their shared left-to-right motion pattern. The key difference lies in the intermediate state: ``vShape'' has a downward intermediate state, while ``inv\_vShape'' has an upward one.

\begin{figure}[h]
\centering 
\subfigure[Training and Testing Accuracy.]{%
\resizebox*{5cm}{!}{\includegraphics{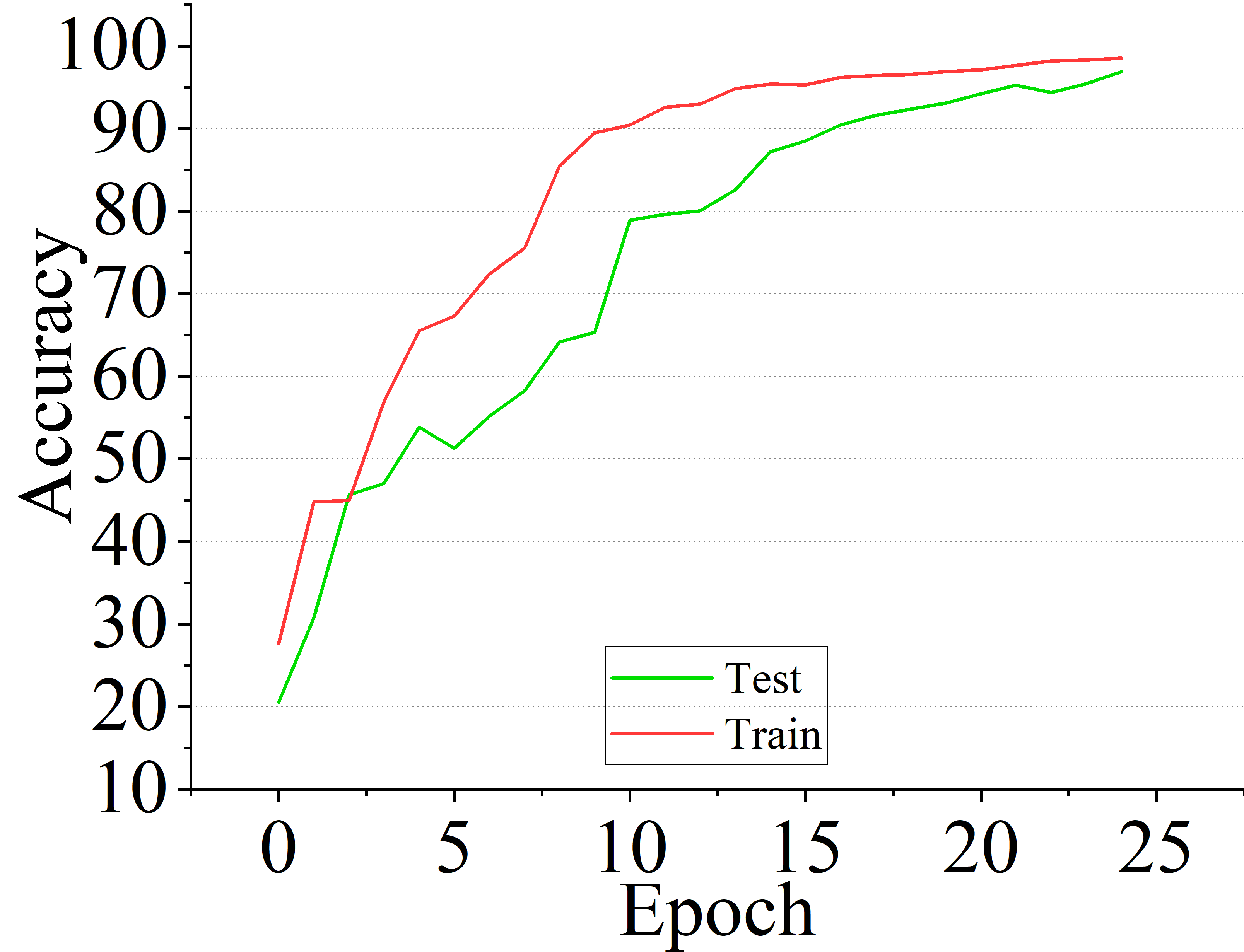}}}
            \hspace{0.01cm}
\subfigure[Training and Testing Loss.]{%
\resizebox*{5cm}{!}{\includegraphics{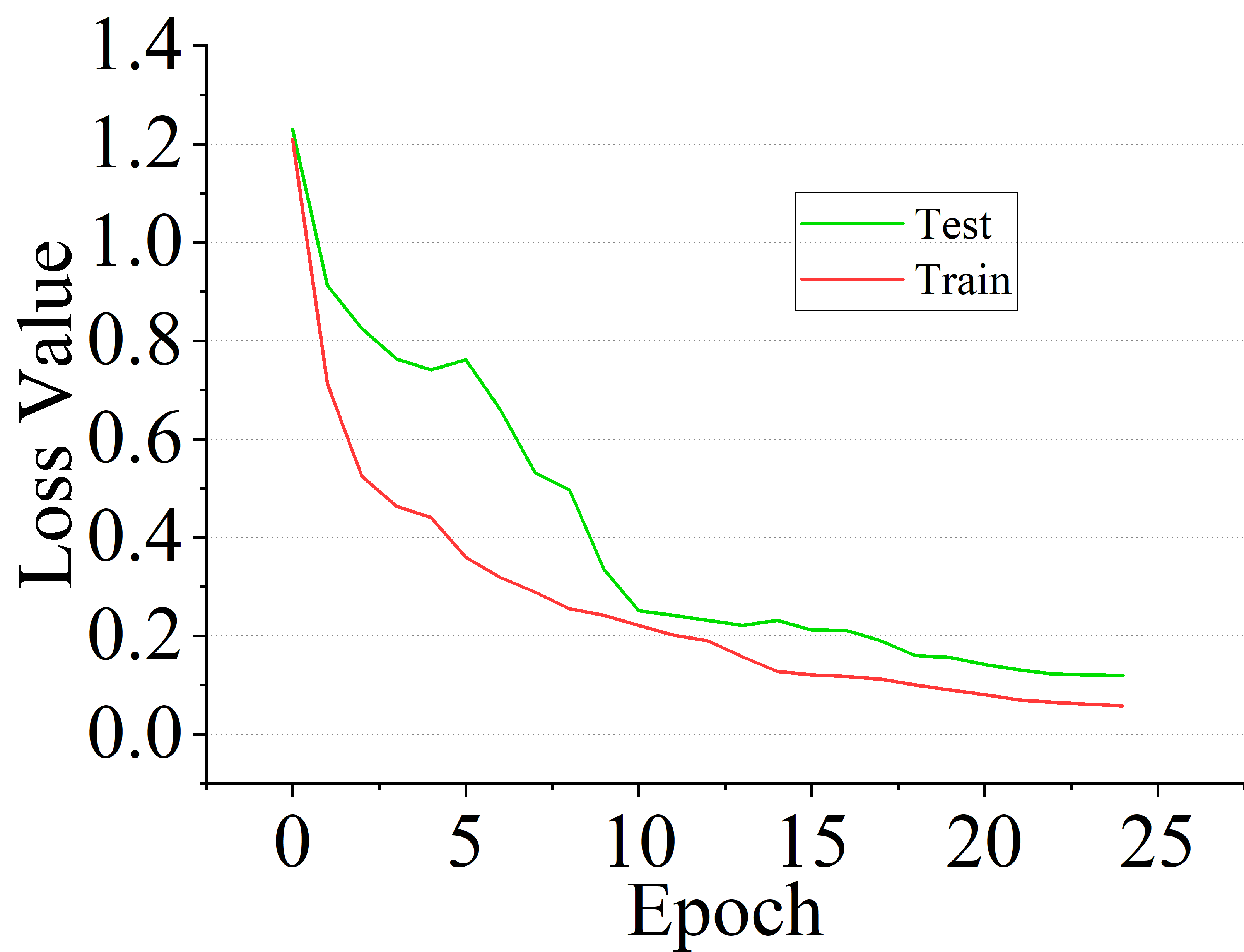}}}
\caption{Evolution of training and testing performance across Epochs.} \label{fig:training_state}
\end{figure}

As for training stability in our model, in Fig. \ref{fig:training_state}, we illustrate the loss and accuracy curves for MAVR-Net on training and testing datasets. The training accuracy (red line in Fig. \ref{fig:training_state}(a)) increases steadily from approximately 20\% to around 99\%, while the test accuracy (red line in Fig. \ref{fig:training_state}(a)) rises more gradually from a similar starting point to about 97\%, demonstrating effective learning. The training loss (red line in Fig. \ref{fig:training_state}(b)) decreases sharply from 1.23 to 0.058, and the test loss (green line in fig. \ref{fig:training_state}(b)) declines with fluctuations to 0.12, indicating a stable training process. The results highlight MAVR-Net's strong learning capability, with a slight performance gap between training and test sets (training slightly higher) and consistent loss differences, suggesting robust training. The curves stabilize at around 23 epochs, indicating minimal benefits from further training. 

Overall, the classification confusion matrix indicates that, despite some instances of MAV actions being misclassified into other categories, MAVR-Net demonstrates robust classification performance for the MAV action class as a whole. Additionally, the trends of training and testing suggest that MAVR-Net exhibits low overfitting risk, high training stability, and strong generalization ability.   

\begin{table*}[h]
    \footnotesize
    \centering
    \caption{Ablation Study Results on Different Configurations.}
    \label{tab:attributeTable}
    \begin{tabular}{ccccccc}
    \hline
    Options & Basic  & MVFPN & Cross                       & Short              & Medium             & Long \\ 
     &   &  &  Attention                      &               &              &  \\ \hline
    Vanilla  & \checkmark &- & -                                 &$87.36\%\pm 8.4\%$  &$86.3\%\pm 7.8\%$   &$84.3\%\pm 13.3\%$  \\ 
    Vanilla+FPN  & \checkmark&\checkmark & -                     &$91.8\%\pm 6.7\%$   &$89.15\%\pm 7.6\%$ &$86.8\%\pm 9.4\%$ \\
    Vanilla+FPN+  &     &    &    & \\
     Attention  &\checkmark & \checkmark&\checkmark   &$97.8\%\pm 2.8\%$   &$96.5\%\pm 3.9\%$   &$92.8\%\pm 4.7\%$ \\
    \hline
    \end{tabular}
\end{table*}

\subsection{Ablation Experiment on MAVR-Net for MAV Action Recognition}
To evaluate the impact of different components of MAVR-Net on MAV action recognition, we conducted ablation experiments on the proposed modules.
First, we analyzed the MVFPN and MV cross-Attention modules separately. 
As summarized in Table \ref{tab:attributeTable}. The baseline model ("vanilla") achieved accuracies of $87.36\%\pm 8.4\%$, $86.3\%\pm 7.8\%$, and $84.3\%\pm 13.3\%$ for Short, Medium, and Long sequences, respectively. Incorporating the MV Feature Pyramid Network (FPN) improved performance to $91.8\%\pm 6.7\%$, $89.15\%\pm 7.6\%$, and $86.8\%\pm 9.4\%$. Further integration of cross-attention ("Vannilla + MV FPN + Attention") yielded the highest accuracies of $97.8\%\pm 2.8\%$, $96.5\%\pm 3.9\%$, and $92.8\%\pm 4.7\%$, demonstrating the significant contributions of both FPN and cross-attention modules. However, compared to MVFPN, the MV Cross-Attention module contributes more significantly to improving MAV action recognition performance. This suggests that MAV action recognition tasks are better suited to neural network models with the ability to extract spatio-temporal contextual information.  

Besides, we also provides the impact of the view alignment loss on the MAV action recognition task. In Table \ref{tab:ali_loss}, the model with the alignment loss (w) achieved accuracies of $97.8\%\pm 2.8\%$, $96.5\%\pm 3.9\%$, and $92.8\%\pm 4.7\%$ for MAV action datasets in Short, Medium, and Long sequences, respectively. In contrast, using the traditional cross-entropy loss without alignment (wo) resulted in lower accuracies of $94.8\%\pm 3.7\%$, $92.84\% \pm 5.6\%$, and $89.8\% \pm 6.4\%$. These findings highlight the effectiveness of the proposed alignment loss in enhancing model performance on our MAV action datasets.

\begin{table}[h]
    \centering
    \caption{Impact of View Alignment Loss on MAV Action Recognition Performance.}
    \label{tab:ali_loss}
    \begin{tabular}{cccc}
    \hline
    Options & Short & Medium & Long \\ \hline
    $w$   &$97.8\%\pm 2.8\%$&$96.5\%\pm 3.9\%$ &$92.8\%\pm 4.7\%$  \\ 
    $wo$  &$94.8\%\pm 3.7\%$&$92.84\%\pm 5.6\%$ &$89.8\%\pm 6.4\%$ \\

    \hline
    \end{tabular}
\end{table}

\subsection{The Impact of Multi-view Experiments}

\begin{figure}[t]
		\centering 
		\includegraphics[width=0.7\textwidth]{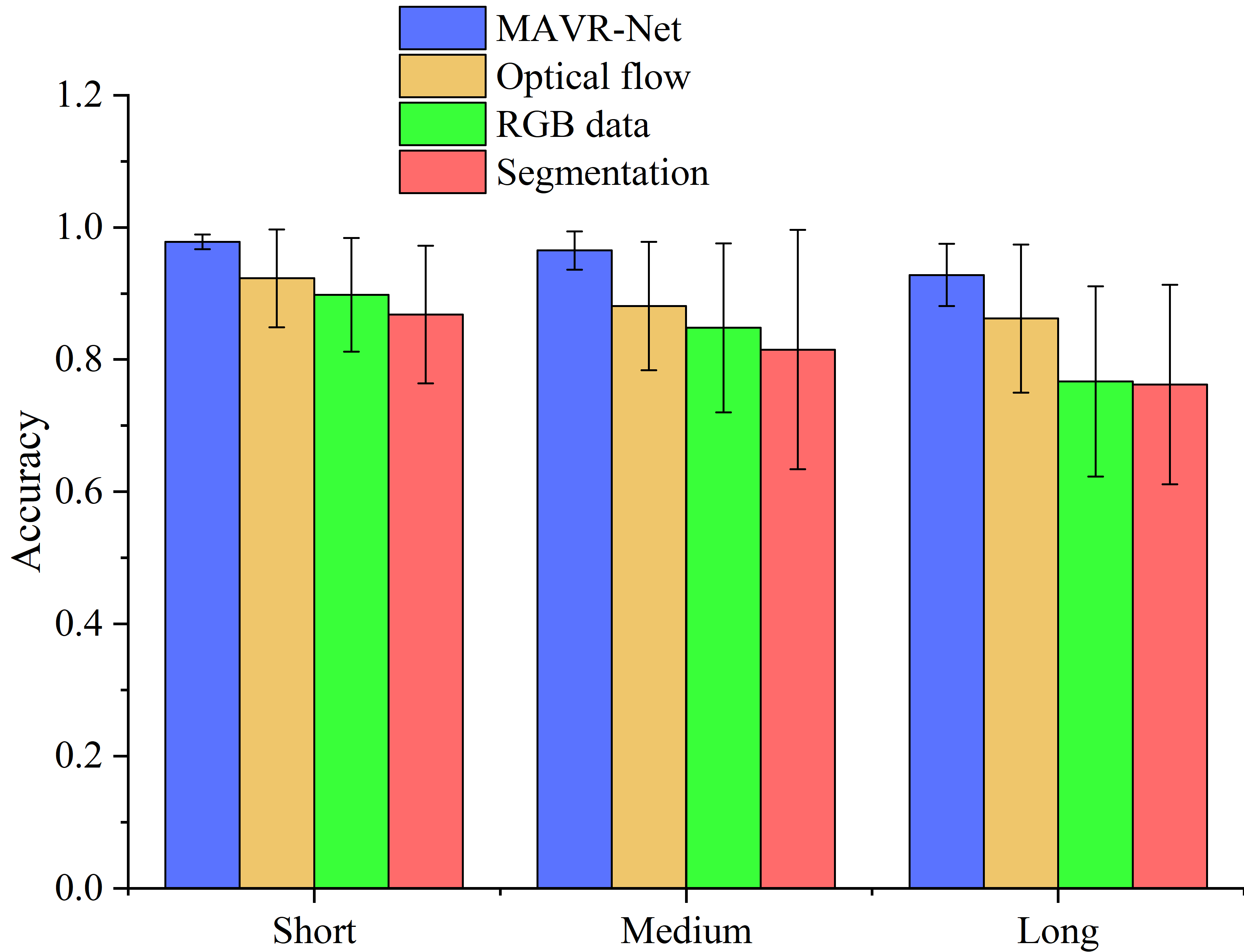}
		\caption{Comparison of MAV action classification performance between MAVR-Net (three-view information) and single-view information.} \label{fig:mvLearning}
\end{figure}

To evaluate the contribution of different view information to MAV action recognition, we conducted multi-view ablation experiments on the MAVR-Net model. During the experiments, models are trained with various view information to observe their impact on MAV action recognition performance. The results of MAV action recognition across different views are shown in Fig. \ref{fig:mvLearning}. Notably, when analyzing single-view information, optical flow achieves the best MAV action classification performance, while segmentation information performed the worst. We attribute this to the fact that optical flow captures significant motion information during MAV movement, providing richer cues for action classification. In contrast, segmentation masks vary with MAV motion, and when the MAV's side faces the RGB sensors, the mask information is significantly reduced, hindering the model's action recognition capability. Furthermore, MAVR-Net with full-view information significantly outperformed single-view models across various action scales. This improvement can be attributed to the complementary information provided by multiple views, enhancing the robustness of MAV action recognition and effectively reducing misclassification rates. These findings confirm that multi-view learning has a positive effect on MAV action recognition, consistent with the conclusions in \cite{yan2021deep} and \cite{chen2022multi}.

\section{Conclusion}
This paper presents a vision-based motion recognition techniques for MAVs, a critical component for achieving visual communication in MAV systems. Specifically, we propose a multi-view learning strategy that extracts distinct view information from MAV motion sequences captured by a visual camera, including optical flow, RGB, and segmentation mask data. A multi-view MAV action recognition neural network, termed MAVR-Net, is developed. This network employs an optical flow branch to encode motion information, an RGB branch to capture appearance information, and a segmentation mask branch to discern shape information, enabling effective recognition of individual MAV motion patterns during flight. To enhance feature integration, a multi-view feature pyramid network fuses multi-stage feature maps from the three view branches, allowing MAVR-Net to learn motion information across different scales. Additionally, a cross-view attention mechanism is introduced to capture complementary information between views, facilitating effective training of the neural network. Experimental results demonstrate that our proposed model outperforms existing action recognition methods in MAV action recognition tasks. Ablation studies further validate the contributions of the multi-view feature pyramid module, cross-view  attention module, and multi-view alignment loss to the overall performance. In the future, we aim to refine the recognition algorithm to efficiently identify MAV action categories in more challenging and resource-constrained environments.

\section*{Acknowledgment}
The authors would like to thank Malaysian Ministry of Higher Education (MOHE) for providing the Fundamental Research Grant Scheme (FRGS) (Grant number: FRGS/1/2024/TK04/USM/02/3) for conducting this research. 






\bibliography{sn-bibliography}

\end{document}